\pdfoutput=1

\documentclass[11pt]{article}

\usepackage[]{EMNLP2022}

\usepackage{times}
\usepackage{latexsym}

\usepackage[T1]{fontenc}

\usepackage[utf8]{inputenc}
\usepackage{microtype}

\usepackage{inconsolata}

\usepackage{times}
\usepackage{latexsym}

\usepackage{graphicx}
\usepackage{colortbl}
\usepackage{bbm}
\usepackage{amsfonts}
\usepackage[normalem]{ulem}
\useunder{\uline}{\ul}{}
\usepackage{amssymb}
\usepackage{amsfonts}
\usepackage{framed}
\usepackage{color}
\usepackage{multicol}
\usepackage{multirow}
\usepackage{makecell}
\usepackage{amsmath}
\usepackage{booktabs}
\usepackage{array}
\usepackage{paralist}
\usepackage{algorithm}
\usepackage{bbm}
\usepackage{dsfont}
\usepackage{adjustbox}
\usepackage{diagbox}
\usepackage{algorithmic}

\AtBeginDocument{\setlength\abovedisplayskip{4pt}}
\AtBeginDocument{\setlength\belowdisplayskip{4pt}}

\title{Robust Question Answering against Distribution Shifts with \\ Test-Time Adaptation: An Empirical Study}

\author{Hai Ye\textsuperscript{$\dag$} \ \ \ Yuyang Ding\textsuperscript{$\ddag$} \ \ \ Juntao Li\textsuperscript{$\ddag$} \ \ \ Hwee Tou Ng\textsuperscript{$\dag$} \\ 
\textsuperscript{$\dag$}Department of Computer Science, National University of Singapore \\
\textsuperscript{$\ddag$}Soochow University, China \\
\texttt{\{yeh,nght\}}\texttt{@comp.nus.edu.sg} \ \
\texttt{yyding.me@gmail.com} \ \  \texttt{ljt@suda.edu.cn}
}
  
\begin{document}
\maketitle
\begin{abstract}

A deployed question answering~(QA) model can easily fail when the test data has a distribution shift compared to the training data. Robustness tuning (RT) methods have been widely studied to enhance model robustness against distribution shifts before model deployment. However, can we improve a model after deployment? To answer this question, we evaluate test-time adaptation~(TTA) to improve a model after deployment. We first introduce \textsc{Cold}QA, a unified evaluation benchmark for robust QA against text corruption and changes in language and domain. We then evaluate previous TTA methods on \textsc{Cold}QA and compare them to RT methods. We also propose a novel TTA method called online imitation learning~(OIL). Through extensive experiments, we find that TTA is comparable to RT methods, and applying TTA after RT can significantly boost the performance on \textsc{Cold}QA. Our proposed OIL improves TTA to be more robust to variation in hyper-parameters and test distributions over time\footnote{Our source code is available at~\url{https://github.com/oceanypt/coldqa-tta}}.


\end{abstract}

\section{Introduction}

How to build a trustworthy NLP system that is robust to distribution shifts is important, since the real world is changing dynamically and a system can easily fail when the test data has a distribution shift compared to the training data~\cite{ribeiro2020beyond,wang2021measure}. Much previous work on robustness evaluation has found model failures on shifted test data. For example, question answering~(QA) models are brittle when dealing with paraphrased questions~\cite{gan-ng-2019-improving}, models for task-oriented dialogues fail to understand corrupted input~\cite{liu-etal-2021-robustness,peng-etal-2021-raddle}, and neural machine translation degrades on noisy text input~\cite{belinkov2017synthetic}. In this work, we study the robustness of QA models to out-of-distribution~(OOD) test-time data.




\begin{figure}[t]
\setlength{\abovecaptionskip}{-0cm}
\setlength{\belowcaptionskip}{-0.5cm}
\begin{center}
\includegraphics[width=6.5cm]{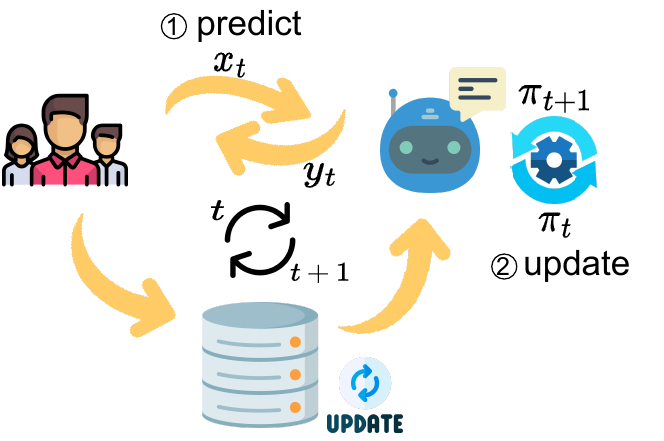}
\end{center}
\caption{Illustration of test-time adaptation. $\pi$ represents the model being adapted. $x_t$ is the test data at time $t$ and $y_t$ is the returned predictions for $x_t$.}
\label{fig:tta-ill}
\end{figure}

To build a model that is robust against distribution shifts, most previous work focuses on robustness tuning~(RT) methods that improve model generalization pre-deployment, such as adversarial training~\cite{madry2017towards}. However, can we continually enhance a model post-deployment? To answer this question, we study and evaluate test-time adaptation~(TTA) for robust QA after model deployment. TTA generalizes a model by continually updating the model with test-time data~\cite{pmlr-v119-sun20b}. As shown in Fig.~\ref{fig:tta-ill}, in this work, we focus on test-time adaptation in real time, where the model predicts and updates over a data stream on the fly. 
For each test data instance, the model first returns its prediction and then updates itself with the test data. 
Unlike unsupervised domain adaptation~\cite{ramponi2020neural} studied in NLP, TTA is suitable for domain generalization, since it makes no assumption about the target distribution and could adapt the model to any arbitrary distribution at test time. 


We discuss TTA methods in $\S$\ref{sec:tta}, where we first present previous popular TTA baselines, and then introduce our newly proposed TTA method, online imitation learning~(OIL). OIL is inspired by imitation learning, where the adapted model learns to clone the actions made by the source model, and the source model aims to reduce overfitting to noisy pseudo-labels in the adapted model. We further adopt causal inference to control model bias from the source model. Next, to compare to TTA methods, we briefly discuss previous robustness tuning~(RT) methods such as adversarial training in $\S$\ref{sec:rt}. 

To study and analyze TTA for robust QA post-deployment, we introduce \textsc{Cold}QA in $\S$\ref{sec:coldqa} which is a unified evaluation benchmark for robust QA against distribution shifts from text corruption, language change, and domain change. It differs from previous benchmarks that only study one type of distribution shifts~\cite{ravichander2021noiseqa,hu2020xtreme,fisch2019mrqa}. 
\textsc{Cold}QA expects a QA model to generalize well to all three types of distribution shifts. 

Our contributions in this work include: 
\begin{compactitem}
 \item We are the first to study test-time adaptation for QA tasks with extensive experiments.
 
  \item We propose a novel TTA method, OIL, which outperforms previous TTA baselines.
   
    \item We propose a new benchmark \textsc{Cold}QA which unifies the evaluation of robust QA against distribution shifts.
   
    \item We evaluate previous robustness tuning methods on the new benchmark. 
\end{compactitem}

\begin{figure}[t]
\setlength{\abovecaptionskip}{-0.1cm}
\setlength{\belowcaptionskip}{-0.5cm}
    \centering
    \begin{minipage}[t]{0.48\columnwidth}
        \centering
        \includegraphics[width=\linewidth]{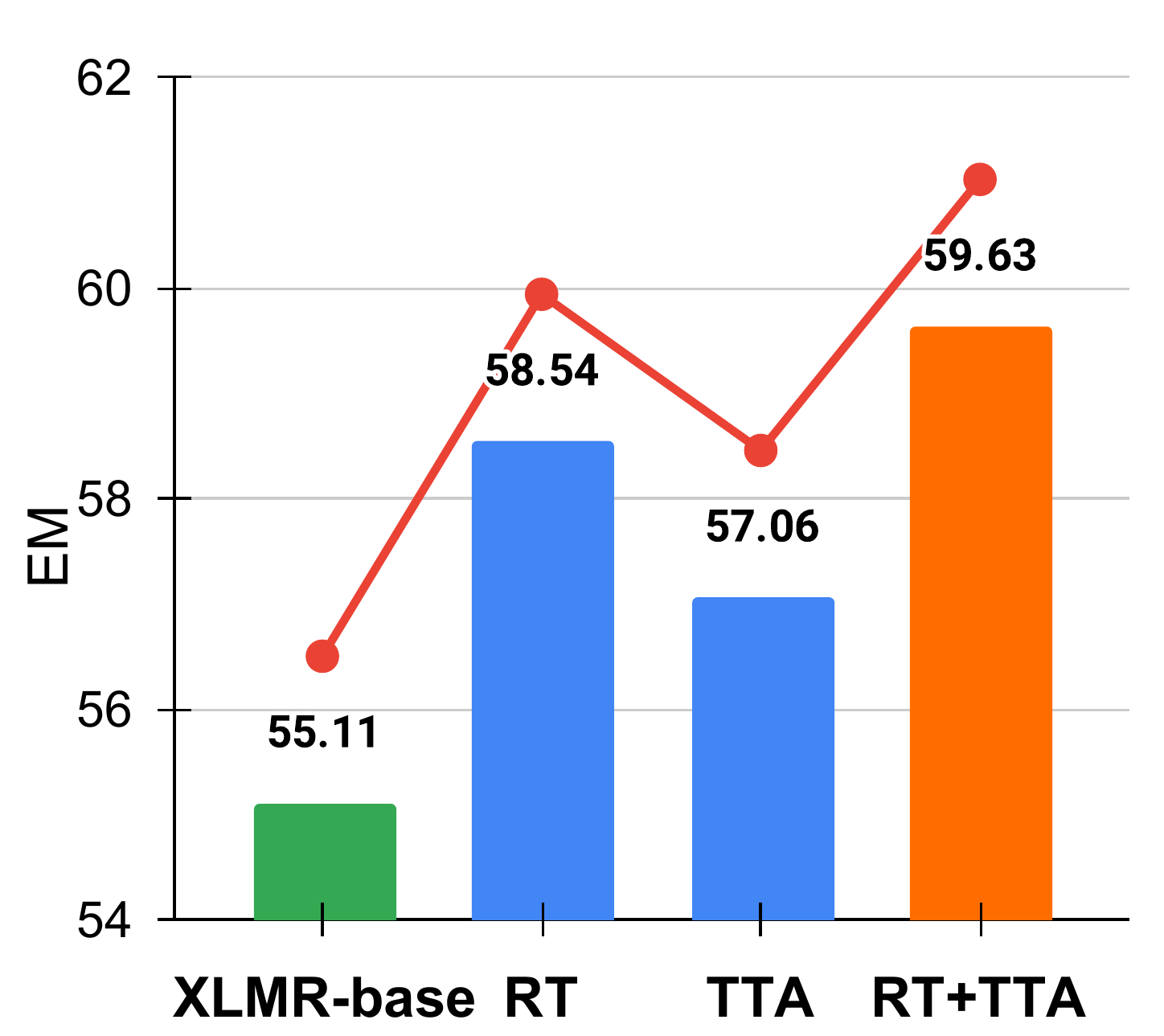}
        \vspace{-0.7em}
    \end{minipage}
    \begin{minipage}[t]{0.48\columnwidth}
        \centering
        \includegraphics[width=\linewidth]{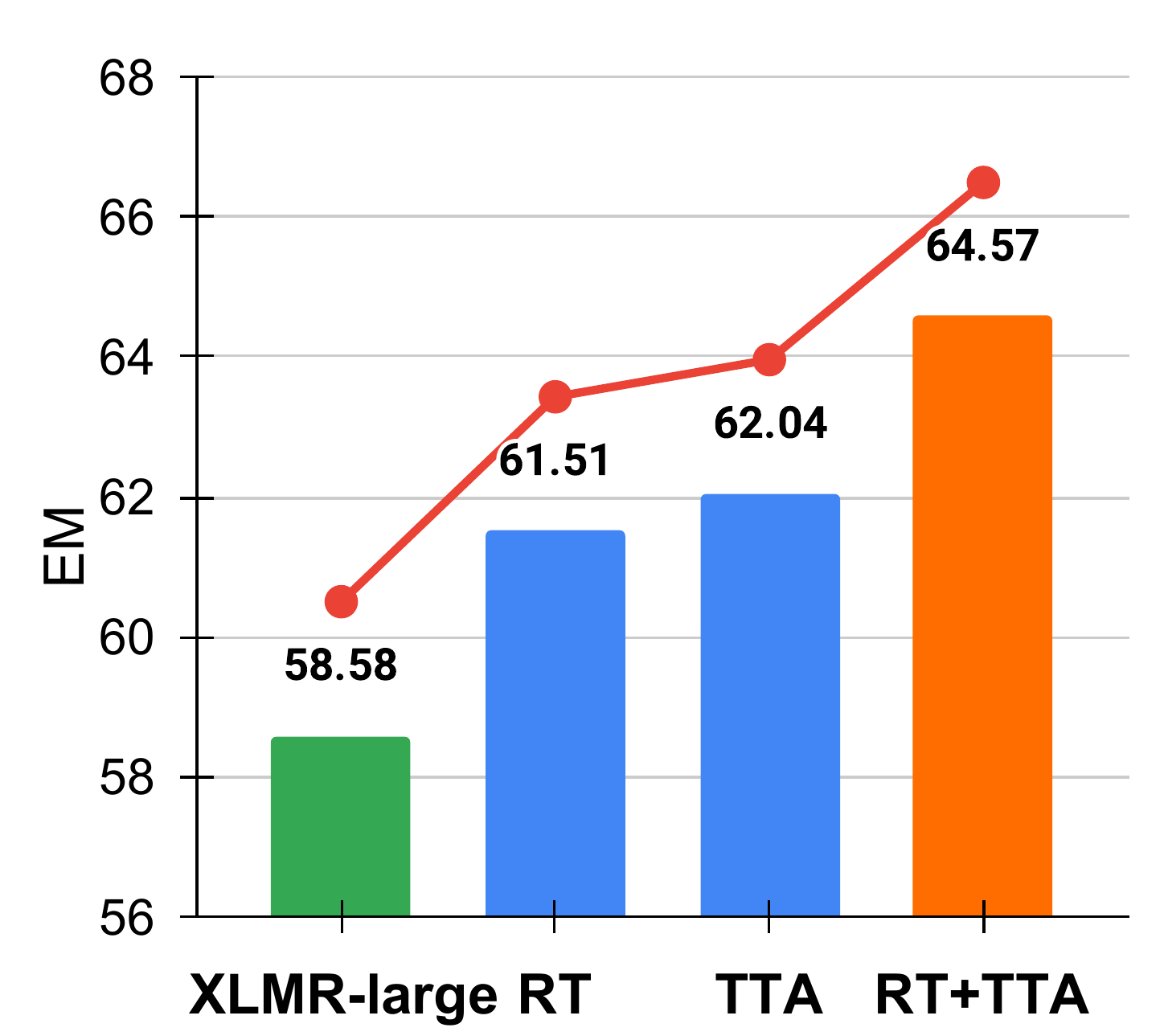}
        \vspace{-0.7em}
    \end{minipage}
 \caption{The average results of RT, TTA, and RT+TTA on \textsc{Cold}QA. RT+TTA significantly improves over RT and TTA.}
\label{fig:imp}
\end{figure}

Based on the experimental results in $\S$\ref{sec:exp}, we report the following findings:
\begin{compactitem}
    \item \textsc{Cold}QA is challenging and not all RT methods are effective on \textsc{Cold}QA~($\S$\ref{sec:main-result});
    \item Overall, as Fig.~\ref{fig:imp} shows, TTA is comparable to RT, and applying TTA after RT can further boost model performance~($\S$\ref{sec:main-result});
    \item Compared to previous TTA baselines, OIL is more robust to changes in hyper-parameters and test distributions over time~($\S$\ref{sec:pl-vs-oil}). 
\end{compactitem}


\begin{table*}
\setlength{\abovecaptionskip}{0.1cm}
\setlength{\belowcaptionskip}{-0.4cm}
\centering
\resizebox{12cm}{!}{%
\begin{tabular}{l|cc|cc}
\toprule
\multirow{2}{*}{\textbf{Settings}} & \multicolumn{2}{c|}{\textbf{Trianing Data}} & \multicolumn{2}{c}{\textbf{Training Loss}} \\ 
                               & \emph{Training time}        & \emph{Test time}  & \emph{Training time}         & \emph{Test time}     \\ \midrule
Unsupervised domain adaptation     & $\mathcal{X}^s$, $\mathcal{Y}^s$; $\mathcal{X}^t$         & None         & $\mathcal{L}(x_s, y_s) + \mathcal{L}(x_t)$        & None       \\ 
Robustness tuning              & $\mathcal{X}^s$, $\mathcal{Y}^s$ & None  & $\mathcal{L}(x^s, y^s)$ & None     \\ \midrule
Test-time adaptation (online) & $\mathcal{X}^s$, $\mathcal{Y}^s$         & $\mathcal{X}^t$  &  None          & $\mathcal{L}(x^t)$ \\ \bottomrule
\end{tabular}%
}
\caption{Compared settings. $\mathcal{X}^s$ and $\mathcal{Y}^s$ are drawn from a source distribution and $\mathcal{X}^t$ from a target distribution. $x^s \in \mathcal{X}^s$, $y^s \in \mathcal{Y}^s$, $x^t \in \mathcal{X}^t$. }
\label{tab:compared-setting}
\end{table*}


\section{Related Work}
\noindent{\textbf{Robust QA}} \ \ Much previous work on model robustness evaluation has shown that NLP models fail on test data with distribution shifts~\cite{rychalska2019models,ribeiro2020beyond,wang2021measure} compared to the training data. For QA tasks, \citet{ravichander2021noiseqa} study how text corruption affects QA performance. \citet{lewis2019mlqa} and \citet{Artetxe:etal:2019} analyze cross-lingual transfer of a QA system. \citet{fisch2019mrqa} benchmark the generalization of QA models to data with domain shift. In this work, we jointly study distribution shifts due to corruption, language change, and domain change. Adversarial samples cause another type of distribution shifts~\cite{jia-liang-2017-adversarial} which is not studied in this work. Hard samples~\cite{ye-etal-2022-robustness}, dataset bias~\cite{tu2020empirical}, and other robustness issues are not the focus of this work. 

\noindent{\textbf{Test-Time Adaptation}} \ \ TTA adapts a source model with test-time data from a target distribution. TTA has been verified to be very effective in image recognition~\cite{pmlr-v119-sun20b,wang2020tent,liu2021ttt++,pmlr-v151-bartler22a}. In NLP, \citet{wang2021efficient} learn to combine adapters on low-resource languages at test time to improve sequence labeling tasks. \citet{gao2022simulating} and \citet{li2022using} keep adapting a QA model after model deployment using user feedback, which is different from our work which requires no human involvement when adapting the model. \citet{ben2022pada} study test-time adaptation for text classification and sequence labeling, but they focus on example-based prompt learning which needs expert knowledge to design prompts. \citet{DBLP:conf/naacl/BanerjeeGB21} explore test-time learning for QA tasks, but their work concerns how to train a QA model from scratch by using unlabeled test data, instead of adapting to out-of-distribution test data.

\noindent{\textbf{Domain Adaptation}} \ \ Different from test-time adaptation, unsupervised domain adaptation~(UDA) needs to know the target domain when performing adaptation pre-deployment~\cite{DBLP:journals/ml/Ben-DavidBCKPV10,DBLP:conf/ijcai/LiHYNB020,DBLP:conf/emnlp/YeTHLNB20,karouzos2021udalm}. UDA tries to minimize the gap between the source and target domain. Recent work studies UDA without knowing the source domain~\cite{liang2020we,su-etal-2022-comparison}, which means the model can be adapted to any unseen target domain on the fly. However, they assume all target data is available when performing adaptation, unlike online adaptation.

\noindent{\textbf{Robustness Tuning}} \ \ Robustness tuning~(RT) is another family of methods that tries to train a more generalized model pre-deployment at training time instead of test time. Adversarial training is a well-studied method to enhance model robustness~\cite{miyato2016adversarial,madry2017towards,zhu2019freelb,wang2020infobert}. Some work also uses regularization to improve model generalization~\cite{wang2021multi,bo2021xtune,cheng2020posterior,jiang2019smart}. Prompt tuning~\cite{DBLP:conf/emnlp/LesterAC21} and adapter-based tuning~\cite{DBLP:conf/acl/HeLYTDCLBS20} can also enhance model generalization to unseen test distributions. 

\noindent{\textbf{Life-long Learning}} \ \ Similar to TTA, life-long learning~(LLL) can also continually improve a model post-deployment~\cite{parisi2019continual}. However, LLL requires the model to remember previously learned knowledge, and training data in the target distribution is labeled. TTA only focuses on the distribution to be adapted to and the test data is unlabeled. \citet{lin-etal-2022-continual} also adapt QA models with test-time data but in a LLL setting.

\section{Test-Time Adaptation}\label{sec:tta}
\noindent{\textbf{Problem Definition}} \ \ Given a source model $\pi_0$ trained on a source distribution $S$, test-time adaptation~(TTA) adapts the model to the test distribution $\mathcal{T}$ with the test data, which enhances the model post-deployment. In the setting of online adaptation, test-time data comes in a stream\footnote{In this work, we do not study offline adaptation.}. As shown in Fig.~\ref{fig:tta-ill}, at time $t$, for the test data $x_t \sim \mathcal{T}$, the model $\pi_t$ will first predict its labels $y_t$ to return to the end user. Next, $\pi_t$ adapts itself with a TTA method and the adapted model will be carried forward to time $t+1$. The process can proceed without stopping as more test data arrive. There is no access to the gold labels of test data in the whole process. 
We compare the setting studied in this work, which is online test-time adaptation, with unsupervised domain adaptation and robustness tuning in Table~\ref{tab:compared-setting}.

\subsection{TTA with Tent and PL}
We first discuss two prior TTA methods, \textbf{Tent}~\cite{wang2020tent} and \textbf{PL}~\cite{lee2013pseudo}. Tent adapts the model by entropy minimization, in which the model predicts the outputs over test-time data and calculates the entropy loss for optimization. Similarly, PL is a pseudo-labeling method, predicting the pseudo-labels on test-time data and calculating the cross-entropy loss. Tent is simple yet it achieves SOTA performance on computer vision (CV) tasks such as image classification, compared to other more complex methods, such as TTT~\cite{pmlr-v119-sun20b} which needs to modify the training process by introducing extra self-supervised losses. Other TTA methods improve over Tent~\cite{pmlr-v151-bartler22a,liu2021ttt++}, but they are much more complex. 

Formally, Tent and PL start from the source model $\pi_0$. At time $t$, the model $\pi_t$ updates itself with the test data $x_t$. The loss for optimization is denoted as $l_t(\pi_t)$:
\begin{equation}
    l_t(\pi_t) =  H(p_t)_{\text{Tent}} \ \ \text{or} \ \ H(p_t, {y}_t)_{\text{PL}}
    \label{eq:tent-pl}
\end{equation}
where $p_t$ is the predicted probabilities over the output classes of $x_t$ from the model $\pi_t$, and ${y}_t = {\arg\max}_{i} {p}_t[i]$. $H(\cdot)$ and $H(,)$ are the entropy and cross-entropy loss respectively. On the data $x_t$, the model is optimized with only one gradient step to get $\pi'_t$: $\pi'_t \leftarrow \pi_t$. Then the model $\pi'_t$ will be carried forward to time $t+1$: $\pi_{t+1} \leftarrow \pi'_t$. 

\subsection{Online Imitation Learning}
Adapting by the model alone, Tent and PL may easily lose the ability to predict correct labels, since the labels predicted by them are not verified to be correct and learning with such noisy signals may degrade the model. The model may not recover again once it starts to deteriorate. To overcome such an issue, inspired by imitation learning~\cite{ross2011reduction}, we propose online imitation learning~(OIL) in this work. OIL aims to train a learner~(or model) $\pi$ by the supervision of an expert $\pi_e$ in a data stream. The expert can help the model to be more robust throughout model adaptation, since the expert is stable and the learner clones the behavior of the expert.

Formally, at each time $t$, the expert $\pi_e$ takes an action~(makes a prediction) $\hat{y}_t \sim \pi_e$ on $x_t \sim \mathcal{T}$. The learner $\pi_t$ then learns to clone such an action by optimizing a surrogate objective $l_t$: 
\begin{equation}
    l_t (\pi_t) = \mathbb{E}_{x_t \sim \mathcal{T}} \mathbb{E}_{\langle y_t, \hat{y}_t \rangle \sim  \langle \pi, \pi_e \rangle } \mathcal{L}(y_t, \hat{y}_t; x_t )
\end{equation}
where $y_t$ is the action taken by the learner at time $t$ and $\mathcal{L}(,;)$ measures the distance between the two actions. 
Formally, at time $T$, with a sequence of online loss functions $\{l_t\}_{t=1}^T$ and the learners $\Pi =  \{\pi_t\}_{t=1}^T$, the regret $R(T)$ is defined as:
\begin{equation}
    R(T) = \sum_{t=1}^{T} l_t (\pi_t) - \min_{\pi \in \Pi}  \sum_{t=1}^T l_t(\pi)
\end{equation}
where we try to minimize such regret during adaptation, which is equal to optimizing the loss function $l_t(\pi_t)$ at each time $t$~\cite{ross2011reduction}. 


\subsubsection{Instantiation of TTA with OIL}
At time $0$, both the learner $\pi$ and the expert $\pi_e$ are initialized by the source model $\pi_0$. 
At time $t$, the loss function $l_t(\pi_t)$ for optimization is:
\begin{equation}
    l_t(\pi_t) = H(p_t, \hat{y}_t) 
    \label{eq:lt}
\end{equation}
where $p_t$ is the predicted probabilities over the output classes of $x_t$ from the learner $\pi_t$. $\hat{y}_t = {\arg\max}_{i} \hat{p}_t[i]$ in which $\hat{p}_t$ is the corresponding predicted probabilities of the expert $\pi_e$. 
Same as Tent and PL, the model is also optimized with one gradient step to get $\pi'_t$: $\pi'_t \leftarrow \pi_t$, and the model $\pi'_t$ is carried forward to time $t+1$: $\pi_{t+1} \leftarrow \pi'_t$. 

For the expert, we can also update it by using the model parameters of the learner. At time $t$, we update the expert as follows:
\begin{equation}
    \theta_{\pi_e} \leftarrow \alpha \cdot \theta_{\pi_e} + (1 - \alpha) \cdot \theta_{\pi'_{t}}
    \label{eq:expert}
\end{equation}
where $\theta$ represents the model parameters and $\alpha$ is a hyper-parameter to control the updating of the expert. $\alpha$ is set to a high value such as 0.99 or 1, so the expert stays close to the source model $\pi_0$ in the adaptation process. Here, the expert is also similar to the mean teacher~\cite{tarvainen2017mean}. 

Furthermore, since the expert is initialized by the source model and because of distribution shift, the actions taken by the expert may be noisy. We can filter and try not to learn these noisy actions. Then the loss function in Eq.~\ref{eq:lt} becomes:
\begin{equation}
    l_t(\pi_t) = \mathbb{I}\big (H(p_t, \hat{y}_t) < \gamma \big ) \cdot H(p_t, \hat{y}_t)
    \label{eq:lt_noise}
\end{equation}
where the cross-entropy loss $H(p_t, \hat{y}_t)$ is used to identify the noisy actions, and $\gamma$ is a hyper-parameter serving as a threshold. 


\begin{figure}[t]
\setlength{\abovecaptionskip}{-0.1cm}
\setlength{\belowcaptionskip}{-0.4cm}
\begin{center}
\includegraphics[width=\columnwidth]{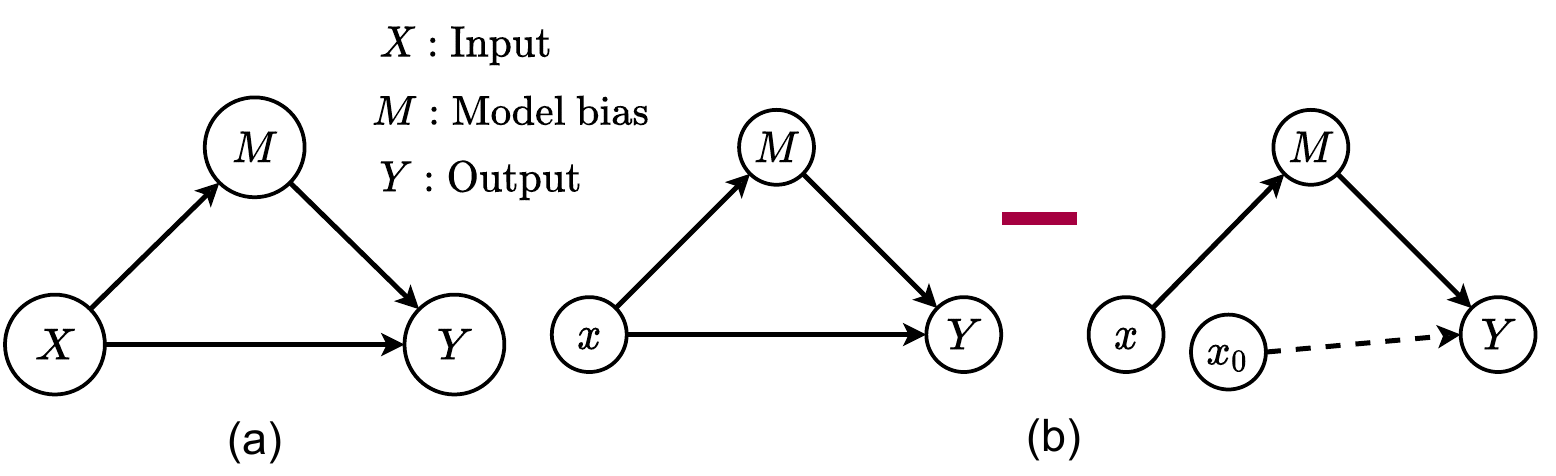}
\end{center}
\caption{(a) The proposed causal graph. (b) The calculation of total direct effect as in Eq.~\ref{eq:tde}.}
\label{fig:causal-graph}
\end{figure}


\subsubsection{Enhancing OIL with Causal Inference}

Since the expert is initialized by the source model, when it predicts labels on the test data, its behavior will be affected by the knowledge that it has learned from the source distribution, which is what we call \emph{model bias} in this work.
Since the test distribution is different from the source distribution, and the expert provides instructions to the learner to clone, such model bias will have a negative effect on the learning of the learner. 
Here, we further use causal inference~\cite{pearl2009causal} to reduce such effect caused by model bias. 

\noindent{\textbf{Causal Graph}} \ \ We assume that the model output of the learner $\pi$ is affected by direct and indirect effect from the input, as shown in the causal graph in Fig.~\ref{fig:causal-graph}a. The causal graph includes the variables which are the input $X$, the output $Y$, and the potential model bias $M$ from the expert. $X \rightarrow Y$ is the direct effect. $X \rightarrow M \rightarrow Y$ represents the indirect effect, where $M$ is a mediator between $X$ and $Y$. $M$ is determined by the input $X$, which can come from in-distribution or out-of-distribution data. 


\noindent{\textbf{Causal Effects}} \ \ Our goal in causal inference is to keep the direct effect but control or remove the indirect effect. As shown in Fig.~\ref{fig:causal-graph}b, we calculate the total direct effect~(TDE) along $X \rightarrow Y$ as follows: 
\begin{align}
    TDE(y) & = Y_{y|do(X=x)} - Y_{y|do(X=x_0)} \nonumber \\ 
    & = Y_{y|X=x} - Y_{y|X=x_0} \label{eq:tde}
\end{align}
where $do$ operation is the causal intervention~\cite{glymour2016causal} which is to remove the confounders to $X$. However, since there is no confounder to $X$ in our assumption, we just omit it. 

\noindent{\textbf{Model Training}} \ \ Given the total direct effect in Eq.~\ref{eq:tde}, we first have to learn the left term $Y_{y|X=x}$ which is the combination of the direct and indirect effect along $X \rightarrow Y$ and $X \rightarrow M \rightarrow Y$ respectively. We use the learner $\pi$ to learn the direct effect. For the indirect effect, the model bias of $M$ exhibits different behaviors to data from different distributions. Since the learner $\pi$ and the expert $\pi_e$ capture the test and source distribution respectively, 
we use the discrepancy in their outputs to represent the model bias. Considering the model bias, the loss function $l_t$ in Eq.~\ref{eq:lt_noise} becomes: 
\begin{equation}
    l_t(\pi_t) = \mathbb{I}\big (H(p_t, \hat{y}_t) < \gamma \big ) \cdot H\big ( p_t + (p_t - \hat{p}_t), \hat{y}_t \big )
    \label{eq:causal}
\end{equation}
where $p_t$ and $\hat{p}_t$ are the predicted probabilities over the output classes of the learner $\pi_t$ and the expert $\pi_e$ respectively. $p_t$ captures the direct effect and $p_t - \hat{p}_t$ learns the indirect effect.



   
   
   

     
     
     

     


    
   

\noindent{\textbf{Inference}} \ \ 
When performing inference, we take the action $y$ which has the largest TDE value. Based on Eq.~\ref{eq:tde} for TDE calculation, we obtain the prediction over the input $x_t$ using the learner $\pi_t$ as:
\begin{align}
    y_{t}  = {\arg \max}_{{i}} TDE({i}) = \ \  \ \ \ \ \ \ \ \ \ \ \ \ \ \ \ \ \ \ \ \ \nonumber  \\ \ \ \ \   \ p_{t}[i]  + (1- \beta) \cdot (p_{t}  - \hat{p}_{t})[i]
    \label{eq:inference}
\end{align}
where $\beta$ controls the contribution of the indirect effect. Here, when calculating the TDE score, we assume the model output is zero when given the null input $x_0$, since we assume the model cannot make predictions without the given input. We set $\beta$ to 1 throughout the experiments, which completely eliminates the effect of model bias. 

\begin{algorithm}[t]
\begin{algorithmic}[1]
   \REQUIRE Source model $\pi_0$; memory bank size $K$; $\alpha$ for expert updating; $\gamma$ for filtering noisy actions; $\beta$ for controlling indirect effect.

   \STATE Initialize the expert $\pi_e \gets \pi_0$;
   
   \FOR{$t = 1, 2, \cdots $}
   
   \STATE Return predictions on $x_t$ using Eq.~\ref{eq:inference};
   
   \STATE Enqueue ${x}_t$ \& dequeue ${x}_{t-K}$;
   
   \FOR{$x_k \in \{x_{t-K}, \cdots, x_t \}$}
     
     \STATE Use $x_k$ to update the learner $\pi_t$ as in Eq.~\ref{eq:causal};
     
    \STATE Update the expert $\pi_e$ as in Eq.~\ref{eq:expert};
     
   \ENDFOR 
    
   \ENDFOR
   
\end{algorithmic}
\caption{Online Imitation Learning}
  \label{alg:OIL}
\end{algorithm}

\subsection{Implementation of TTA for the QA Task}
For extractive question answering, the model needs to predict the start and end position. The above TTA methods treat the two positions independently and apply the same loss, i.e., $l_t(\pi_t)$, to them separately, and the final loss takes the average of the two. We present the pseudocode of OIL in Algorithm~\ref{alg:OIL}, where Tent and PL follow the same procedure but with different losses to update. 
The data $x_t$ at each time $t$ is a batch of instances. We preserve a memory bank with size $K$ to store the data from time $t-K$ to $t$, which more fully exploits test-time data for model adaptation. At each time $t$, we enqueue $x_t$ and dequeue $x_{t-K}$ from the memory bank. Then each batch of data from the memory bank is used to optimize the online loss as shown in Eq.~\ref{eq:causal}. The expert for OIL is updated accordingly.

\section{Robustness Tuning}\label{sec:rt}

In contrast to improving the model post-deployment with TTA, robustness tuning~(RT) enhances the model pre-deployment. RT has been studied in NLP to improve model generalization~\cite{wang2021measure}. 
RT methods are applied at training time when training the source model. 
We also benchmark RT methods on \textsc{Cold}QA to compare with TTA methods.

First, we compare with adversarial training methods, which are \textbf{FGM}~\cite{miyato2016adversarial}, \textbf{PGD}~\cite{madry2017towards}, \textbf{FreeLB}~\cite{zhu2019freelb}, and \textbf{InfoBERT}~\cite{wang2020infobert}. 
Next, we further evaluate robustness tuning methods proposed for cross-lingual transfer, which are \textbf{MVR}~\cite{wang2021multi} and \textbf{xTune}~\cite{bo2021xtune}. These two methods use regularization to enhance model robustness. All of these methods have not been comprehensively evaluated on distribution shifts arising from text corruption, language change, and domain change. 

\noindent{\textbf{Combination of RT and TTA.}} \ \ Finally, we also study combining RT and TTA methods. The source model is tuned by a RT method, then this model is adapted by a TTA method to the test distribution. 

\begin{table*}[t]
\setlength{\abovecaptionskip}{-0cm}
\setlength{\belowcaptionskip}{-0.0cm}
\centering
\small
\resizebox{14cm}{!}{%
\begin{tabular}{lccllccc}
\toprule[.8pt]
\textbf{Source} &
  \textbf{$|$Train$|$} &
  \textbf{$|$Dev$|$} &
  \textbf{Distribution Shift} &
  \textbf{Target} &
  \textbf{$|$Subset$|$} &
  \textbf{$|$Test$|$} &
  \textbf{Metric} \\ \midrule[.5pt]
\multirow{9}{*}{SQuAD} &
  \multirow{9}{*}{87,599} &
  \multirow{9}{*}{34,726} &
  \multirow{2}{*}{Text corruption} &
  NoiseQA-syn &
  3 &
  1,190 &
  \multirow{9}{*}{EM / F1} \\
 &  &  &                                  & NoiseQA-na & 3  & 1,190        &  \\ \cline{4-7} 
 &  &  & \multirow{2}{*}{Language change} & XQuAD      & 11 & 1,190        &  \\
 &  &  &                                  & MLQA       & 7  & 4,517–11,590 &  \\ \cline{4-7}
 &  &  & \multirow{5}{*}{Domain change}   & HotpotQA   & 1  & 5,901        &  \\ 
 &  &  &                                  & NaturalQA  & 1  & 12,836       &  \\ 
 &  &  &                                  & NewsQA     & 1  & 4,212        &  \\ 
 &  &  &                                  & SearchQA   & 1  & 16,980       &  \\
 &  &  &                                  & TriviaQA   & 1  & 7,785        &  \\ \toprule[.8pt]
\end{tabular}%
}
\caption{Detailed characteristics of \textsc{Cold}QA. To perform evaluation on \textsc{Cold}QA, a model is first trained on the source distribution. Next, the trained model is tested on each subset of each target dataset.}
\label{tab:coldQA}
\end{table*}

\begin{table*}
\setlength{\abovecaptionskip}{0.1cm}
\setlength{\belowcaptionskip}{-0.2cm}
\centering
\resizebox{\textwidth}{!}{%
\begin{tabular}{l|l|ll|llllllllll}
\toprule
 &
   &
  \multicolumn{2}{l|}{\textbf{ColdQA}} &
  \multicolumn{4}{l}{\textbf{Text Corruption}} &
  \multicolumn{4}{l}{\textbf{Language Change}} &
  \multicolumn{2}{l}{\textbf{Domain Change}} \\ \midrule
 &
   &
  \multicolumn{2}{l|}{Average} &
  \multicolumn{2}{l}{NoiseQA-syn} &
  \multicolumn{2}{l}{NoiseQA-na} &
  \multicolumn{2}{l}{XQuAD} &
  \multicolumn{2}{l}{MLQA} &
  \multicolumn{2}{l}{MRQA} \\ \midrule
 &
  Metric &
  EM &
  F1 &
  EM &
  F1 &
  EM &
  F1 &
  EM &
  F1 &
  EM &
  F1 &
  EM &
  F1 \\ \midrule
 &
  xlmr-base &
  55.11 &
  69.21 &
  66.64 &
  78.67 &
  66.05 &
  77.91 &
  55.59 &
  71.42 &
  47.14 &
  65.27 &
  40.11 &
  52.78 \\ \midrule
\multirow{2}{*}{\textbf{RT}} &
  MVR &
  56.93 &
  70.51 &
  68.85 &
  80.10 &
  67.87 &
  78.91 &
  58.08 &
  73.34 &
  48.45 &
  66.33 &
  41.40 &
  53.84 \\
 &
  xTune &
  58.54 &
  71.94 &
  70.95 &
  81.52 &
  69.75 &
  80.66 &
  58.78 &
  73.75 &
  49.87 &
  67.76 &
  43.36 &
  56.03 \\ \midrule
\multirow{3}{*}{\textbf{TTA}} &
  Tent &
  56.24 &
  69.68 &
  68.02 &
  79.37 &
  67.79 &
  78.91 &
  57.40 &
  72.56 &
  47.59 &
  65.13 &
  40.39 &
  52.43 \\
 &
  PL &
  56.45 &
  69.78 &
  68.51 &
  79.59 &
  68.15 &
  79.23 &
  57.91 &
  72.69 &
  47.75 &
  65.17 &
  39.94 &
  52.20 \\
 &
  OIL &
  57.06 &
  70.38 &
  68.75 &
  79.86 &
  68.40 &
  79.40 &
  57.96 &
  72.64 &
  48.39 &
  66.08 &
  41.80 &
  53.92 \\ \midrule
\multirow{2}{*}{\textbf{RT+TTA}} &
  xTune + PL &
  58.86 &
  71.89 &
  71.73 &
  82.12 &
  \textbf{70.87} &
  81.10 &
  \textbf{60.23} &
  \textbf{74.56} &
  50.33 &
  68.10 &
  41.14 &
  53.58 \\
 &
  xTune + OIL &
  \textbf{59.63} &
  \textbf{72.68} &
  \textbf{71.90} &
  \textbf{82.24} &
  70.81 &
  \textbf{81.15} &
  60.13 &
  74.46 &
  \textbf{50.67} &
  \textbf{68.53} &
  \textbf{44.65} &
  \textbf{57.00} \\ \midrule \midrule
 &
  xlmr-large &
  58.58 &
  73.82 &
  65.55 &
  79.91 &
  64.17 &
  78.37 &
  63.15 &
  78.77 &
  53.87 &
  72.58 &
  46.18 &
  59.46 \\ \midrule
\multirow{6}{*}{\textbf{RT}} &
  FGM &
  58.06 &
  73.56 &
  64.93 &
  79.71 &
  62.94 &
  77.96 &
  63.21 &
  78.74 &
  54.14 &
  72.72 &
  45.09 &
  58.65 \\
 &
  PGD &
  58.80 &
  74.02 &
  65.91 &
  80.16 &
  63.75 &
  78.05 &
  63.80 &
  78.91 &
  54.28 &
  72.82 &
  46.25 &
  60.19 \\
 &
  FreeLB &
  58.79 &
  73.83 &
  66.22 &
  79.97 &
  64.37 &
  77.88 &
  63.34 &
  78.79 &
  53.93 &
  72.51 &
  46.07 &
  59.99 \\
 &
  InfoBERT &
  57.72 &
  73.39 &
  64.59 &
  79.52 &
  62.66 &
  77.46 &
  62.31 &
  78.28 &
  53.98 &
  72.52 &
  45.05 &
  59.14 \\
 &
  MVR &
  59.52 &
  74.51 &
  67.06 &
  80.96 &
  64.76 &
  78.59 &
  63.35 &
  78.61 &
  54.47 &
  72.80 &
  47.97 &
  61.60 \\
 &
  xTune &
  61.51 &
  76.06 &
  70.11 &
  83.17 &
  67.20 &
  80.54 &
  65.00 &
  79.91 &
  \textbf{56.30} &
  74.33 &
  48.94 &
  62.37 \\ \midrule
\multirow{3}{*}{\textbf{TTA}} &
  Tent &
  54.56 &
  70.34 &
  52.91 &
  69.01 &
  54.29 &
  69.87 &
  63.22 &
  78.91 &
  52.72 &
  70.96 &
  49.65 &
  62.95 \\
 &
  PL &
  61.80 &
  76.05 &
  71.26 &
  83.60 &
  69.32 &
  81.67 &
  64.05 &
  79.21 &
  54.27 &
  72.57 &
  50.12 &
  63.21 \\
 &
  OIL &
  62.04 &
  76.19 &
  71.57 &
  83.93 &
  70.11 &
  82.22 &
  64.19 &
  79.37 &
  54.41 &
  72.90 &
  49.93 &
  62.53 \\ \midrule
\multirow{2}{*}{\textbf{RT+TTA}} &
  xTune + PL &
  63.73 &
  77.01 &
  76.01 &
  86.60 &
  \textbf{73.83} &
  84.55 &
  65.74 &
  \textbf{80.15} &
  55.78 &
  73.92 &
  47.29 &
  59.81 \\ 
 &
  xTune + OIL &
  \textbf{64.57} &
  \textbf{77.93} &
  \textbf{76.13} &
  \textbf{86.72} &
  73.69 &
  \textbf{84.61} &
  \textbf{65.83} &
  80.12 &
  56.24 &
  \textbf{74.34} &
  \textbf{51.00} &
  \textbf{63.86} \\ \bottomrule
\end{tabular}%
}
\caption{Benchmarking results~(\%) on \textsc{Cold}QA for XLMR-base and XLMR-large. Each TTA method is run three times with random seeds and the average results are reported. \textbf{Bold}: the best results. 
}
\label{tab:main-result}
\end{table*}

\begin{table*}[t]
\setlength{\abovecaptionskip}{0.1cm}
\setlength{\belowcaptionskip}{-0.2cm}
\centering
\small
\resizebox{14cm}{!}{%
\begin{tabular}{l|ccccc|c}
\toprule
\textbf{MRQA} & \textbf{HotpotQA}      & \textbf{NaturalQA} & \textbf{NewsQA}        & \textbf{TriviaQA}      & \textbf{SearchQA} & \textbf{Average}          \\ \midrule
xlmr-large & 53.13 / 68.19 & 44.38 / 61.26          & 45.56 / 63.35 & 58.25 / 67.16 & 29.56 / 37.34          & 46.18 / 59.46 \\  \midrule
xTune      & 55.41 / 70.85 & 47.44 / 63.39          & \underline{47.98} / \underline{65.58} & \underline{60.95} / \underline{70.19} & 32.92 / 41.82          & 48.94 / 62.37 \\  \midrule
Tent       & 53.95 / 69.20 & 48.14 / 64.34          & 45.61 / 63.36 & 58.95 / 67.77 & \textbf{41.61 / 50.09}          & 49.65 / 62.95 \\ 
PL         & 53.83 / 69.03 & 50.88 / 66.17          & 46.05 / 63.68 & 58.56 / 67.31 & \underline{41.26} / \underline{49.84} & \underline{50.12} / \underline{63.21} \\ 
OIL        & \underline{56.65} / \underline{71.92} 
 & \textbf{54.49 / 68.16} & 46.77 / 64.12 & 59.18 / 67.99 & 32.54 / 40.45          & 49.93 / 62.53 
 \\ \midrule
xTune+PL      & 55.95 / 71.30          & 53.90 / \underline{68.11}      & \textbf{48.50 / 65.72} & 59.75 / 68.78          & 18.33 / 25.13     & 47.29 / 59.81          \\ 
xTune+OIL     & \textbf{58.46 / 73.99} & \underline{54.27} / 68.06      & 47.45 / 64.80          & \textbf{61.05 / 70.21} & 33.74 / 42.23     & \textbf{51.00 / 63.86} \\ \bottomrule
\end{tabular}%
}
\caption{Results~(EM / F1) on each subset of MRQA. \textbf{Bold}: the best results. \underline{Underlined}: the second best results.}
\label{tab:subset-mrqa}
\end{table*}

\section{\textsc{Cold}QA}\label{sec:coldqa}
To study robust QA under distribution shifts, in this work we introduce \textsc{Cold}QA, a unified evaluation benchmark against text \underline{co}rruption, \underline{l}anguage change, and \underline{d}omain change. As shown in Table~\ref{tab:coldQA}, we collect some existing QA datasets to construct the source and target distributions for \textsc{Cold}QA.

\noindent{\textbf{Source Distribution}} \ \ The training data for the source distribution is SQuAD v1.1~\cite{rajpurkar2016squad}. To evaluate model generalization on \textsc{Cold}QA, we first need to train a source model with the source training data. Next, we evaluate the model on each subset of each target dataset. For test-time adaptation, the model needs to be adapted with the test data on the fly. 
To evaluate performance under all kinds of distribution shifts, we use a multilingual pre-trained language model as the base model since it maps different languages into a shared representation space. 

\noindent{\textbf{Target Distributions}} \ \  We study the following target distribution shifts at test time.

$\bullet$ \textbf{Text Corruption} \ We use NoiseQA to evaluate model robustness to text corruption. NoiseQA~\cite{ravichander2021noiseqa} studies noises from real-world interfaces, i.e., speech recognizers, keyboards, and translation systems. When humans use these interfaces, the questions asked may contain noises, which degrade the QA system's performance. NoiseQA includes two subsets, NoiseQA-na and NoiseQA-syn. NoiseQA-na has real-world noises annotated by human annotators, while NoiseQA-syn is synthetically generated. 


$\bullet$ \textbf{Language Change} \ A robust QA system should also perform well when the inputs are in other languages. 
We use the datasets XQuAD \cite{Artetxe:etal:2019} and MLQA \cite{lewis2019mlqa}, designed for cross-lingual transfer, to evaluate change of language in the test data. 

$\bullet$ \textbf{Domain Change} \ The test data may come from a domain different from the source domain used for model training. Here, the training and test domains are in the same language without any text corruption. We use the datasets from MRQA \cite{fisch2019mrqa} for evaluation, which include HotpotQA~\cite{DBLP:conf/emnlp/Yang0ZBCSM18}, NaturalQA~\cite{47761}, NewsQA~\cite{DBLP:conf/rep4nlp/TrischlerWYHSBS17}, SearchQA~\cite{DBLP:journals/corr/DunnSHGCC17}, and TriviaQA~\cite{DBLP:conf/acl/JoshiCWZ17}. The development sets of these datasets are used.  

\noindent{\textbf{Comparison to Existing Benchmarks}} \ \ To the best of our knowledge, \textsc{Cold}QA is the first benchmark that unifies robustness evaluation over text corruption, language change, and domain change. Previous benchmarks for robust QA usually only study one type of these distribution shifts, e.g., NoiseQA~\cite{ravichander2021noiseqa}, XTREME~\cite{hu2020xtreme}, and MRQA~\cite{fisch2019mrqa} study text corruption, language change, and domain change respectively, where the methods proposed on these benchmarks are tested only on one type of distribution shifts. So it is unclear if prior proposed methods generalize well to other types of distribution shifts. In contrast, \textsc{Cold}QA evaluates a method on all types of distribution shifts mentioned above, a more challenging task to tackle.

\section{Experiments}\label{sec:exp}

\subsection{Setup}
To carry out comprehensive evaluation on all types of distribution shifts, we use a multilingual pre-trained language model as the base model, specifically XLMR-base and XLMR-large~\cite{conneau2019unsupervised}. 
To train the source model on SQuAD with vanilla fine-tuning, we use the default training setup from \citet{hu2020xtreme}. For robustness tuning, we use the hyper-parameter values suggested by \citet{wang2020infobert} to train FreeLB and InfoBERT. For MVR and xTune, the default settings from the original work are used\footnote{We implemented FGM, PGD, FreeLB, and InfoBERT ourselves, since there is no open-source code that can be directly used.}. 


For test-time adaptation, the details of setting the hyper-parameter values for learning rate, batch size, $\alpha$, $\gamma$, and $K$ are given in Appendix A and shown in Table ~\ref{tab:hyper-para}. 
All model parameters are updated during adaptation. Dropout is turned off for Tent and PL when generating the model outputs or pseudo-labels.
The adaptation time of OIL on each dataset from \textsc{Cold}QA is shown in Table~\ref{tab:eval-time-mrqa}, \ref{tab:eval-time-mlqa}, and \ref{tab:eval-time-xquad-noiseqa}. All experiments were performed on one NVIDIA A100 GPU.

\begin{figure*}[t]
\setlength{\abovecaptionskip}{-0.1cm}
\setlength{\belowcaptionskip}{-0.1cm}
    \centering
    \begin{minipage}[t]{0.32\textwidth}
        \centering
        \includegraphics[width=\linewidth]{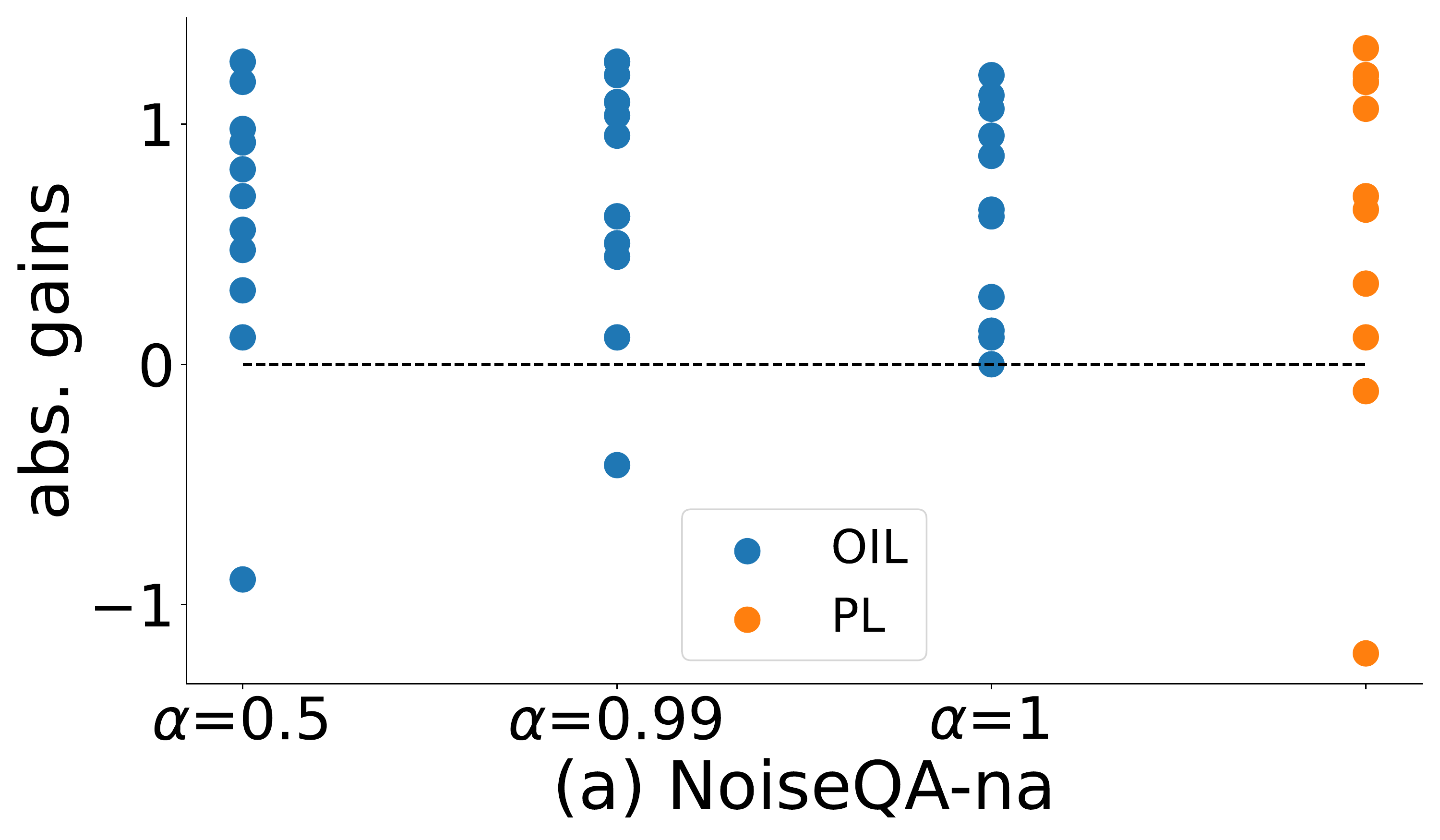}
        \vspace{-0.7em}
    \end{minipage}
    \begin{minipage}[t]{0.32\textwidth}
        \centering
        \includegraphics[width=\linewidth]{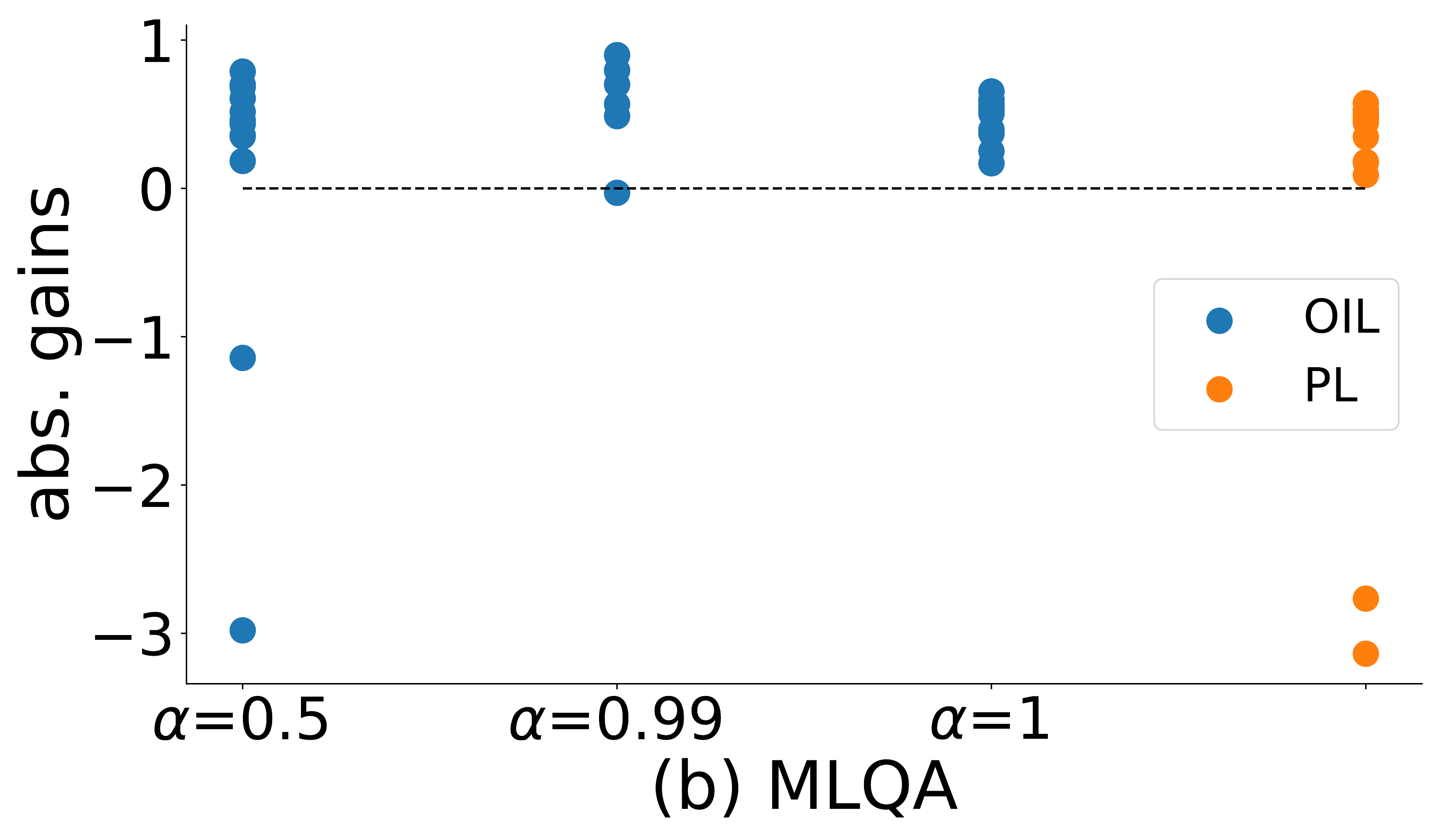}
        \vspace{-0.7em}
    \end{minipage}
    \begin{minipage}[t]{0.32\textwidth}
        \centering
        \includegraphics[width=\linewidth]{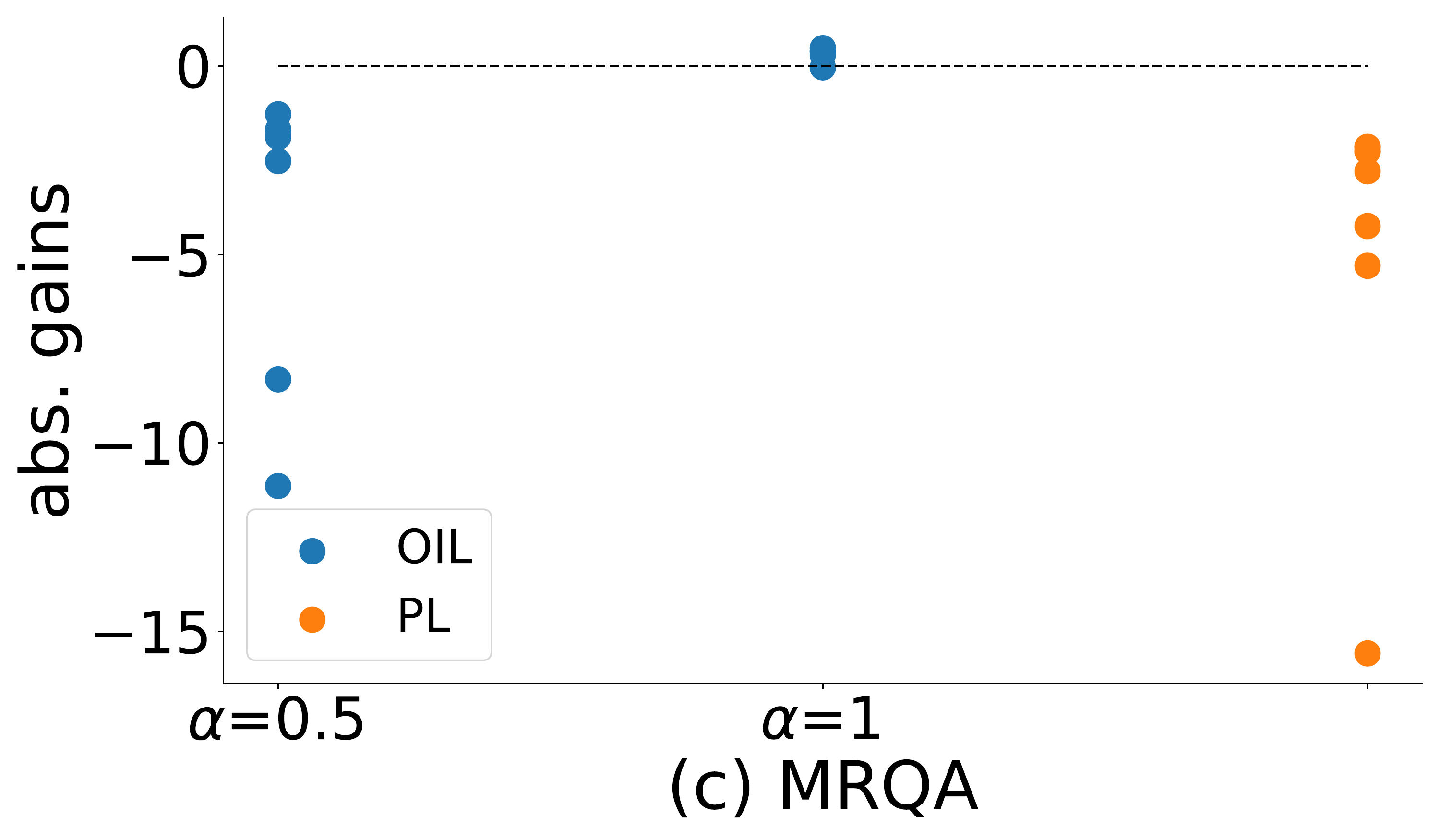}
        \vspace{-0.7em}
    \end{minipage}
\caption{Robustness of PL and OIL to variance of hyper-parameter values. With OIL and PL, we adapt XLMR-base tuned by xTune by using various value combinations of hyper-parameters, which are learning rate and memory size. Learning rate is selected from \{5e-6, 1e-6, 5e-7, 1e-7\} and memory size from \{1, 3, 5\}. For OIL, different values of $\alpha$ are tested.}
\label{fig:oil-analysis}
\end{figure*}

\subsection{Main Results}\label{sec:main-result}
Table~\ref{tab:main-result} shows the benchmarking results of TTA, RT, and their combination on \textsc{Cold}QA. The detailed results on each subset of MRQA are reported in Table~\ref{tab:subset-mrqa}. We have the following observations.

\noindent{\textbf{\textsc{Cold}QA is challenging on which not all RT methods are effective.}} \ \ 
In Fig.~\ref{fig:gain-squad-dev} from the appendix, we report the gains of RT baselines over vanilla fine-tuning on the development set of SQuAD. Not surprisingly, each RT baseline improves the model results on the in-distribution set. However, after re-benchmarking the RT baselines on \textsc{Cold}QA, we see that xTune and MVR are more effective than the adversarial training baselines. Among the adversarial training methods, only PGD and FreeLB can improve the average results but the improvements are marginal. Overall, \textsc{Cold}QA introduces new challenges to the existing RT methods. 

\noindent{\textbf{OIL is stronger than PL and Tent.}} \ \ Tent is much less effective than OIL and PL on \textsc{Cold}QA though it is a very strong baseline on CV tasks~\cite{wang2020tent}. This shows the necessity of re-analyzing TTA methods on QA tasks. OIL is consistently better than PL based on the average results in Table~\ref{tab:main-result}. OIL mostly outperforms Tent and PL based on the detailed results of MRQA in Table~\ref{tab:subset-mrqa}. 

\noindent{\textbf{TTA and RT are both effective and they are comparable to each other.}} \ \ On XLMR-base and XLMR-large, both TTA~(OIL and PL) and RT~(xTune and MVR) can significantly improve the average results by around 1-3 absolute points. Overall, TTA and RT are comparable to each other. More specifically, on XLMR-large, the best TTA method which is OIL outperforms xTune on the average results. On XLMR-large, OIL is better than xTune on NoiseQA and MRQA, but lags behind xTune on XQuAD and MLQA. However, on XLMR-base, TTA does not outperform RT. We think the reason is that the effectiveness of TTA depends on the source model, since TTA starts from the source model and the source model decides the accuracy of predicted pseudo-labels on the test data. 

\noindent{\textbf{Applying TTA after RT significantly boosts the performance and achieves SOTA results.}} On both XLMR-base and XLMR-large, xTune+OIL achieves the best average performance compared to all other methods. 
On XLMR-large, xTune+OIL improves xTune by 3 points on EM score. Among the three types of distribution shifts, xTune+OIL is more effective on text corruption and domain change than language change. Finally, xTune+OIL improves over the baseline EM score by more than 4 points on XLMR-base and 6 points on XLMR-large, significantly improving QA robustness against distribution shifts. 





\begin{table*}[t]
\setlength{\abovecaptionskip}{0.1cm}
\setlength{\belowcaptionskip}{-0.2cm}
\centering
\small
\resizebox{14cm}{!}{%
\begin{tabular}{lc|ccccc}
\toprule
EM / F1 & Avg.          & NoiseQA-syn   & NoiseQA-na    & XQuAD         & MLQA          & MRQA          \\ \midrule
OIL     & 57.06 / 70.38 & 68.75 / 79.86 & 68.40 / 79.40 & 57.96 / 72.64 & 48.39 / 66.08 & 41.80 / 53.92 \\
\ \ \ \ w/o CI  & 56.68 / 69.85 & 68.41 / 79.48 & 68.11 / 79.14 & 57.98 / 72.43 & 48.08 / 65.65 & 40.82 / 52.57 \\
xTune + OIL & 59.63 / 72.68 & 71.90 / 82.24 & 70.81 / 81.15 & 60.13 / 74.46 & 50.67 / 68.53 & 44.65 / 57.00 \\
\ \ \ \ w/o CI  & 59.06 / 72.11 & 71.13 / 81.76 & 70.18 / 80.64 & 59.73 / 73.93 & 50.49 / 68.31 & 43.75 / 55.92 \\ \bottomrule
\end{tabular}%
}
\caption{Effects of causal inference on XLMR-base. CI: causal inference}
\label{tab:cl}
\end{table*}

\begin{figure}[t]
\setlength{\abovecaptionskip}{-0.1cm}
\setlength{\belowcaptionskip}{-0.3cm}
\begin{center}
\includegraphics[width=\columnwidth]{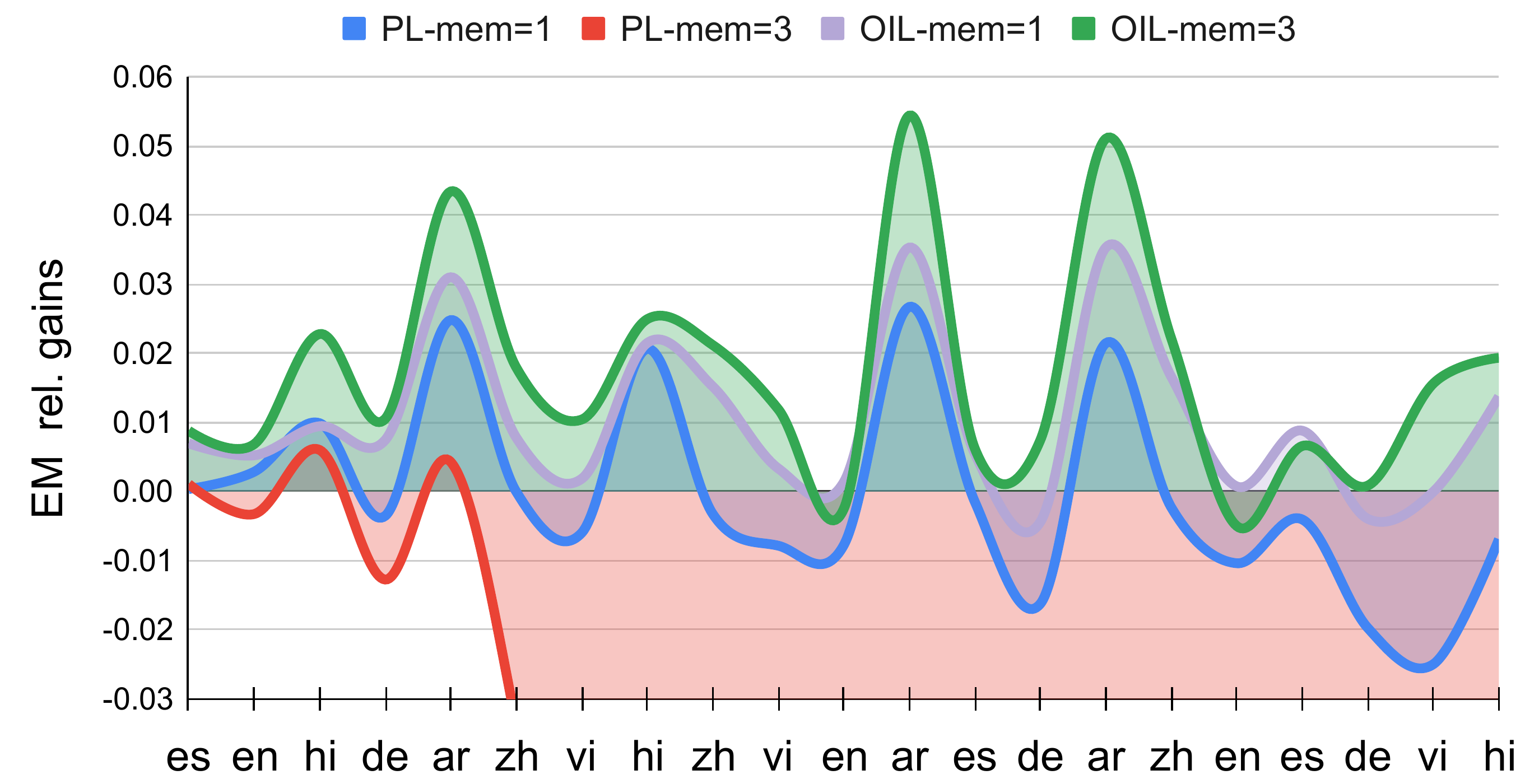}
\end{center}
\caption{Robustness of PL and OIL to changes of test distribution over time. $y$-axis shows relative gains on each language over XLMR-base tuned by xTune. $x$-axis shows test sets from MLQA. mem: memory size $K$. OIL sets $\alpha$ to 0.999.}
\label{fig:ca}
\end{figure}

\subsection{Further Analysis}\label{sec:pl-vs-oil}

\noindent{\textbf{Compared to PL, OIL is more robust to varying hyper-parameter values.}} \ \ OIL utilizes an expert model to perform model adaptation. $\alpha$ controls updating of the expert model. In Fig.~\ref{fig:oil-analysis}, we fix the value of $\alpha$, adapt the model with various combinations of hyper-parameter values, and report the absolute gains on EM score after adaptation. We observe that for OIL, a higher $\alpha$ value such as 0.99 or 1 achieves positive gains under varying hyper-parameter values. However, PL is less stable than OIL under varying hyper-parameter values. To be robust to varying hyper-parameter values is important for TTA, since tuning hyper-parameters on unknown test data is difficult. 

\noindent{\textbf{OIL is better than PL when dealing with changes in test distribution.}} \ \ We further evaluate TTA methods in the setting of continual adaptation, where the test distribution changes over time (rather than staying fixed as in Table~\ref{tab:main-result}), and the model needs to be adapted continually without stopping. In Fig.~\ref{fig:ca}, we adapt the source model from the test language of es to the language hi without stopping. On each test distribution, we report the relative gain over the source model without adaptation. We find that PL is less robust in such a setting and often has negative gains, especially in the last few adaptations. However, our proposed method OIL achieves positive gains among nearly all adaptations, which demonstrates the robustness of OIL in continual adaptation.

\begin{figure}[t]
\setlength{\abovecaptionskip}{0.cm}
\setlength{\belowcaptionskip}{-0.3cm}
    \centering
    \begin{minipage}[t]{0.45\columnwidth}
        \centering
        \includegraphics[width=0.9\linewidth]{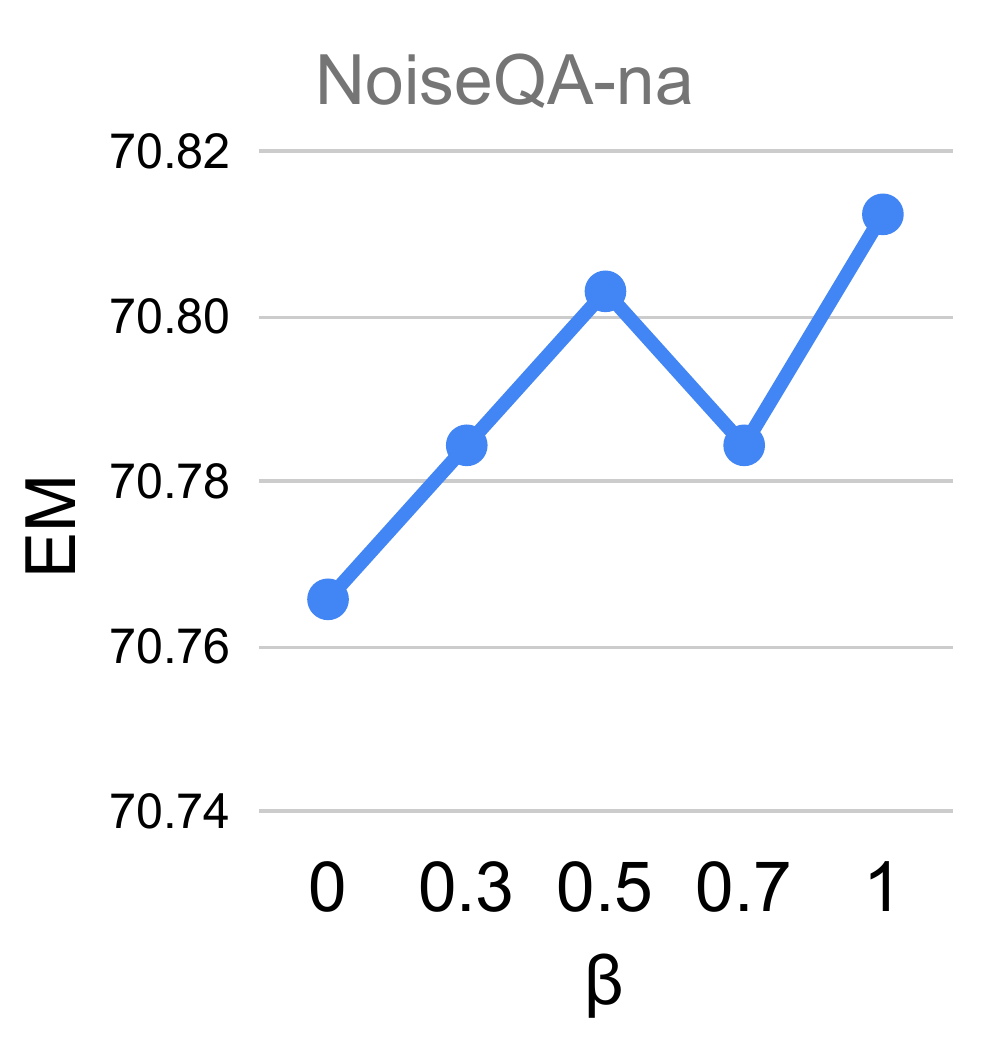}
    \end{minipage}
    \begin{minipage}[t]{0.45\columnwidth}
        \centering
        \includegraphics[width=0.9\linewidth]{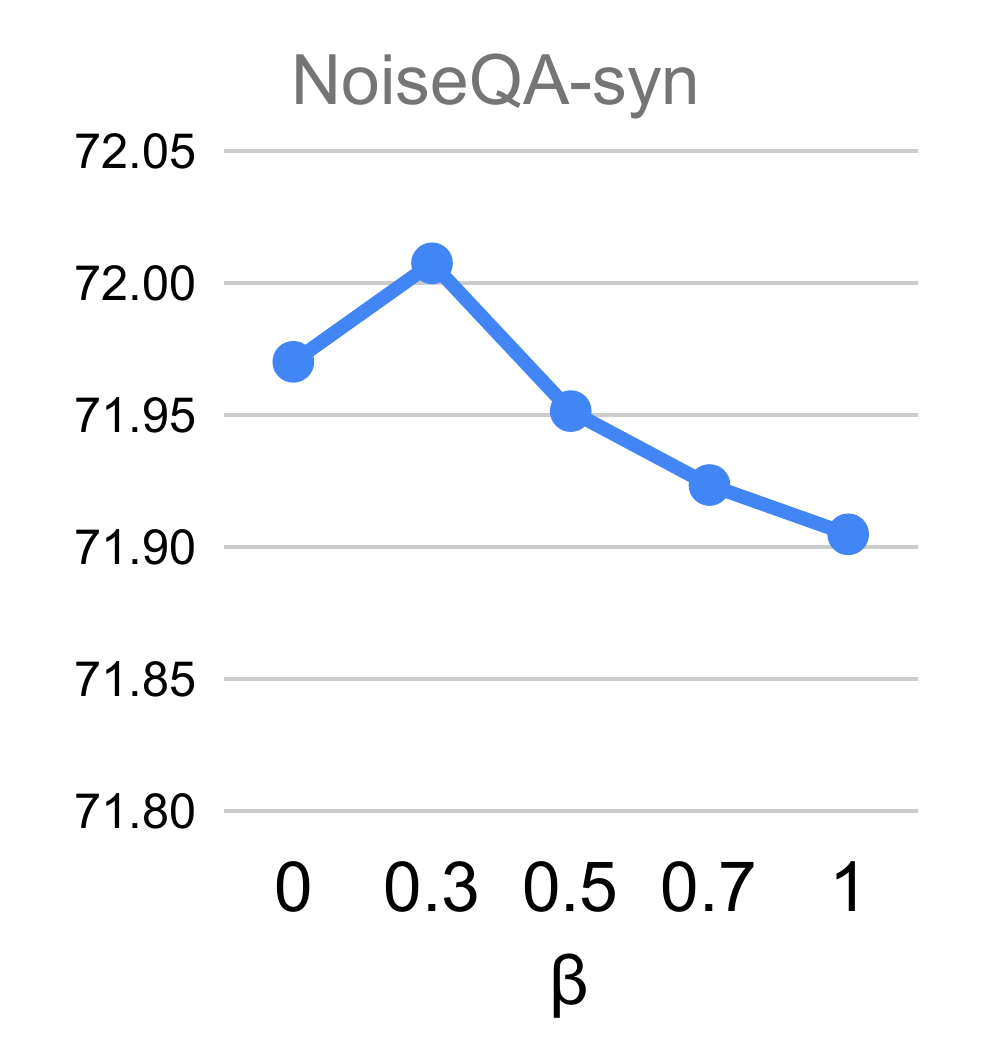}
    \end{minipage}
 \caption{The effects of $\beta$ used in causal inference. Results are evaluated on NoiseQA with XLMR-base tuned by xTune.}
\label{fig:beta}
\end{figure}

\noindent{\textbf{Effects of Causal Inference.}} \ \ Table~\ref{tab:cl} shows the effects of removing causal inference in OIL. Without causal inference, adaptation performance consistently drops on the test sets. In Fig.~\ref{fig:beta}, we further show how $\beta$ affects causal inference. $\beta$ does affect the final results and the optimal $\beta$ value varies with different datasets. To avoid tuning $\beta$, model bias is completely removed by setting $\beta$ to 1 in all our experiments.

\section{Conclusion}
We study test-time adaptation (TTA) for robust question answering under distribution shifts. A unified evaluation benchmark, \textsc{Cold}QA, over text corruption, language change, and domain change is provided. A novel TTA method, OIL, is proposed that achieves good performance when combined with a robustness tuning method.

\section*{Acknowledgements}
This research is supported by the National Research Foundation, Singapore under its AI Singapore Programme (AISG Award No: AISG-RP-2018-007 and AISG2-PhD-2021-08-016[T]). The computational work for this article was partially
performed on resources of the National Supercomputing Centre, Singapore (https://www.nscc.sg).

\section*{Limitations}
Though test-time adaptation shows strong improvements for robust QA under distribution shifts, it still has some issues that need to be addressed in the future. First, model updating is costly. TTA needs to update the model online. However, the cost of updating should be controlled especially for large pre-trained language models. Second, how to choose suitable hyper-parameter values for adaptation is also important. The test data is usually not available and we cannot tune the hyper-parameters before adaptation, so how to effectively select hyper-parameter values is important. In our work, we did not perform hyper-parameter search for OIL. We have also demonstrated in Fig~\ref{fig:oil-analysis} that OIL is robust to various combinations of hyper-parameter values with the help of the expert model. 


\bibliography{anthology,custom}
\bibliographystyle{acl_natbib}

\newpage

\appendix

\section{Appendix}

\begin{table}[t]
\centering
\resizebox{\columnwidth}{!}{%
\begin{tabular}{lccccc}
\hline
OIL    & HotpotQA & NaturalQA & NewsQA & SearchQA & TriviaQA \\ \hline
xlmr-base   & 247      & 771       & 502    & 1265     & 579      \\
xlmr-large  & 1116     & 3450      & 1441   & 6779     & 4369     \\ \hline
\end{tabular}%
}
\caption{Adaptation time~(in seconds) on MRQA.}
\label{tab:eval-time-mrqa}
\resizebox{6cm}{!}{%
\begin{tabular}{lccccccc}
\hline
       & \multicolumn{7}{c}{MLQA}                 \\ \hline
OIL    & en   & es  & de  & ar  & hi  & vi  & zh  \\ \hline
xlmr-base  & 489  & 159 & 164 & 218 & 203 & 239 & 172 \\
xlmr-large & 1635 & 533 & 548 & 724 & 679 & 802 & 574 \\ \hline
\end{tabular}
}
\caption{Adaptation time~(in seconds) on MLQA.}
\label{tab:eval-time-mlqa}
\resizebox{3.5cm}{!}{
\begin{tabular}{lcc}
\hline
 OIL      & XQuAD & NoiseQA \\ \hline
xlmr-base  & 124   & 162     \\
xlmr-large & 187   & 193     \\ \hline
\end{tabular}
}
\caption{Adaptation time~(in seconds) on each subset of XQuAD and NoiseQA.}
\label{tab:eval-time-xquad-noiseqa}
\end{table}

\begin{table}[t]
\centering
\small
\resizebox{6cm}{!}{%
\begin{tabular}{|l|cc|cc|}
\hline
\textbf{NoiseQA-na}  & \multicolumn{2}{c|}{\textbf{OIL}} & \multicolumn{2}{c|}{\textbf{w/o denoise}} \\ \hline
XLMR-base   & \multicolumn{1}{c|}{EM}    & F1    & \multicolumn{1}{c|}{EM}    & F1    \\ \hline
asr         & \multicolumn{1}{c|}{63.59} & 74.74 & \multicolumn{1}{c|}{62.35} & 74.05 \\ 
keyboard    & \multicolumn{1}{c|}{72.04} & 82.90 & \multicolumn{1}{c|}{72.18} & 83.21 \\ 
translation & \multicolumn{1}{c|}{69.58} & 80.55 & \multicolumn{1}{c|}{70.08} & 81.18 \\ 
avg.        & \multicolumn{1}{c|}{68.40} & 79.40 & \multicolumn{1}{c|}{68.21} & 79.48 \\ \hline
XLMR-large  & \multicolumn{1}{c|}{EM}    & F1    & \multicolumn{1}{c|}{EM}    & F1    \\ \hline
asr         & \multicolumn{1}{c|}{62.10} & 75.34 & \multicolumn{1}{c|}{47.98} & 66.36 \\ 
keyboard    & \multicolumn{1}{c|}{74.96} & 86.61 & \multicolumn{1}{c|}{74.43} & 86.24 \\ 
translation & \multicolumn{1}{c|}{73.28} & 84.72 & \multicolumn{1}{c|}{73.14} & 84.70 \\ 
avg.        & \multicolumn{1}{c|}{70.11} & 82.22 & \multicolumn{1}{c|}{65.18} & 79.10 \\ \hline
\textbf{NoiseQA-syn} & \multicolumn{2}{c|}{\textbf{OIL}} & \multicolumn{2}{c|}{\textbf{w/o denoise}} \\ \hline
XLMR-base   & \multicolumn{1}{c|}{EM}    & F1    & \multicolumn{1}{c|}{EM}    & F1    \\ \hline
asr         & \multicolumn{1}{c|}{70.14} & 81.97 & \multicolumn{1}{c|}{69.55} & 81.59 \\ 
keyboard    & \multicolumn{1}{c|}{66.97} & 77.29 & \multicolumn{1}{c|}{66.72} & 77.41 \\ 
translation & \multicolumn{1}{c|}{69.13} & 80.31 & \multicolumn{1}{c|}{69.16} & 80.19 \\ 
avg.        & \multicolumn{1}{c|}{68.75} & 79.86 & \multicolumn{1}{c|}{68.48} & 79.73 \\ \hline
\textbf{XQuAD}       & \multicolumn{2}{c|}{\textbf{OIL}} & \multicolumn{2}{c|}{\textbf{w/o denoise}} \\ \hline
XLMR-base   & \multicolumn{1}{c|}{EM}    & F1    & \multicolumn{1}{c|}{EM}    & F1    \\ \hline
en          & \multicolumn{1}{c|}{72.83} & 83.75 & \multicolumn{1}{c|}{72.91} & 83.89 \\ 
es          & \multicolumn{1}{c|}{60.14} & 77.37 & \multicolumn{1}{c|}{59.55} & 77.33 \\ 
de          & \multicolumn{1}{c|}{58.18} & 74.25 & \multicolumn{1}{c|}{58.63} & 74.48 \\ 
el          & \multicolumn{1}{c|}{56.58} & 73.05 & \multicolumn{1}{c|}{56.13} & 73.15 \\ 
ru          & \multicolumn{1}{c|}{58.26} & 74.28 & \multicolumn{1}{c|}{58.18} & 74.51 \\ 
tr          & \multicolumn{1}{c|}{52.38} & 67.89 & \multicolumn{1}{c|}{51.57} & 68.01 \\ 
ar          & \multicolumn{1}{c|}{49.89} & 66.40 & \multicolumn{1}{c|}{49.83} & 66.72 \\ 
vi          & \multicolumn{1}{c|}{54.99} & 73.83 & \multicolumn{1}{c|}{55.41} & 74.11 \\ 
th          & \multicolumn{1}{c|}{63.59} & 72.22 & \multicolumn{1}{c|}{62.21} & 71.58 \\ 
zh          & \multicolumn{1}{c|}{59.66} & 68.59 & \multicolumn{1}{c|}{58.15} & 67.85 \\ 
hi          & \multicolumn{1}{c|}{51.04} & 67.38 & \multicolumn{1}{c|}{51.09} & 68.17 \\ 
avg.        & \multicolumn{1}{c|}{57.96} & 72.64 & \multicolumn{1}{c|}{57.61} & 72.71 \\ \hline
\end{tabular}%
}
\caption{Effects of not filtering noisy labels in OIL for NoiseQA and XQuAD.}
\label{tab:denoise}
\end{table}

\begin{figure}[t]
\setlength{\abovecaptionskip}{-0.1cm}
\setlength{\belowcaptionskip}{-0.5cm}
\begin{center}
\includegraphics[width=5.5cm]{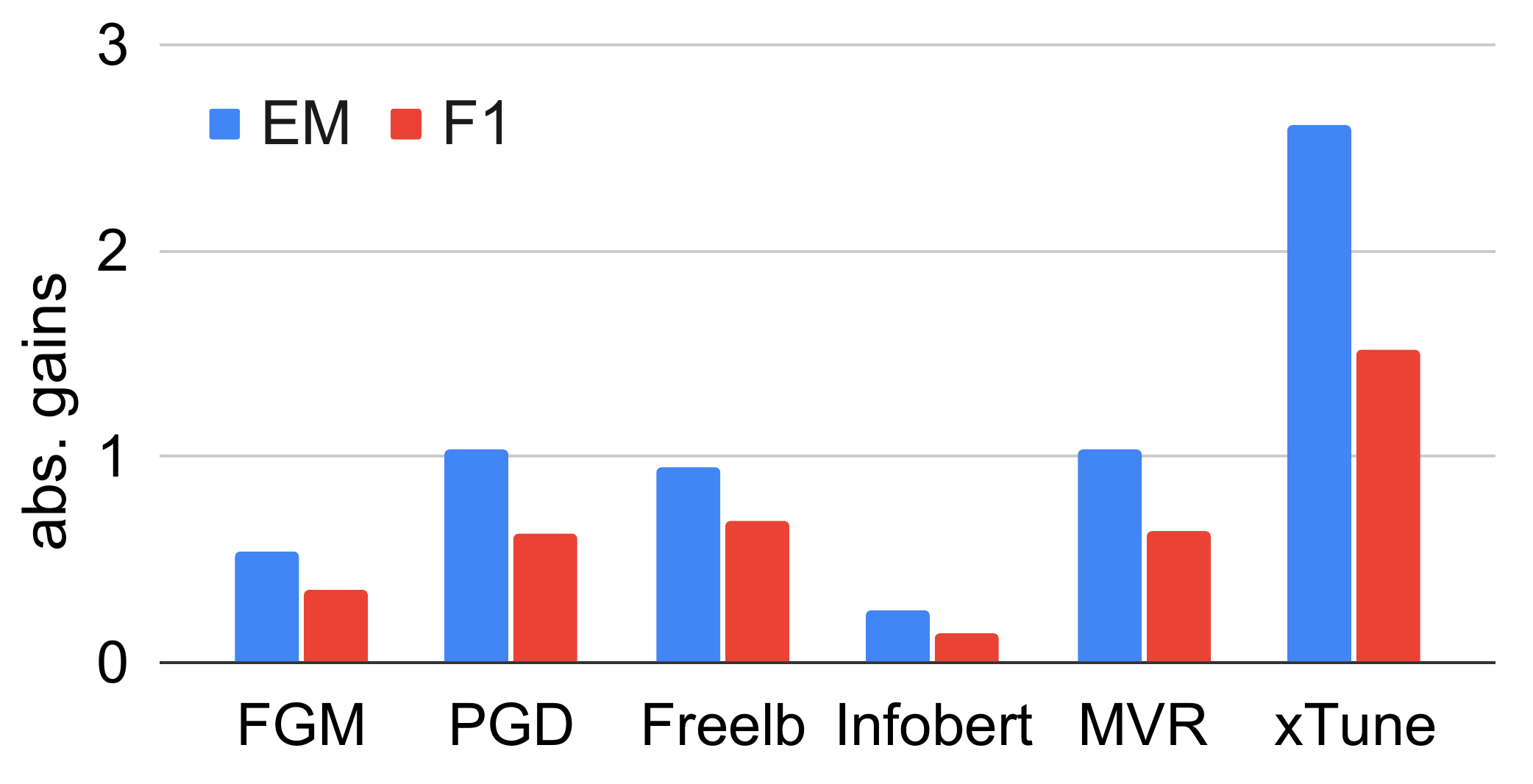}
\end{center}
\caption{Gains over the results of XLMR-large with vanilla fine-tuning on the development set of SQuAD. abs.: absolute.}
\label{fig:gain-squad-dev}
\end{figure}

\begin{figure}[t]
\setlength{\abovecaptionskip}{-0cm}
\setlength{\belowcaptionskip}{-0.5cm}
\begin{center}
\includegraphics[width=\columnwidth]{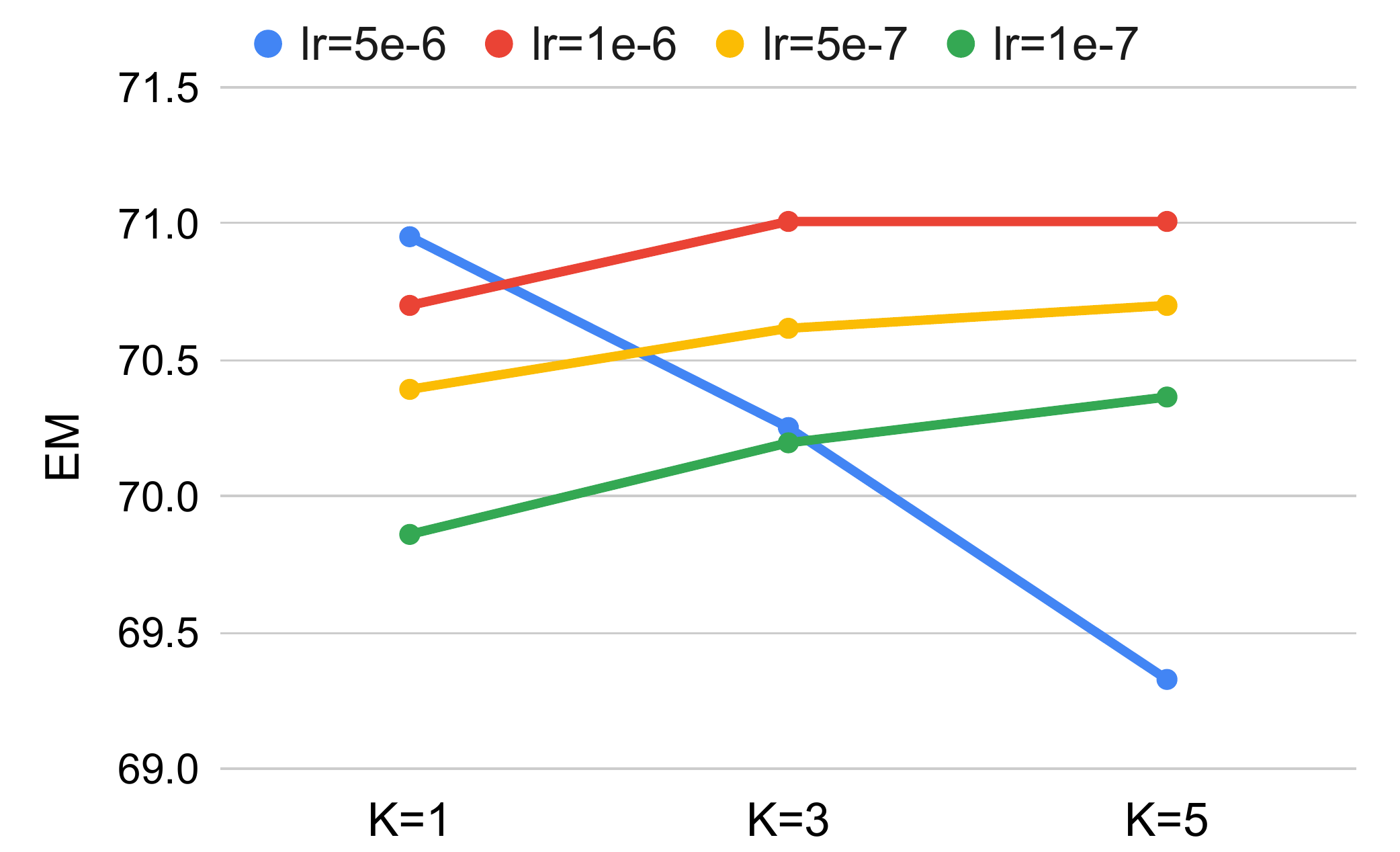}
\end{center}
\caption{Effects of memory size $K$ on NoiseQA-na. The source model is XLMR-base tuned by xTune.}
\label{fig:memory-size}
\end{figure}

\begin{table*}[t]
\centering
\resizebox{\textwidth}{!}{%
\begin{tabular}{l|c|c|c|c|c|c|c|c}
\toprule
\textbf{Dataset} &
  \textbf{NoiseQA} &
  \textbf{XQuAD} &
  \textbf{MLQA} &
  \textbf{HotpotQA} &
  \textbf{NaturalQA} &
  \textbf{NewsQA} &
  \textbf{SearchQA} &
  \textbf{TriviaQA} \\ \midrule
\textbf{|Size|} &
  1,190 &
  1,190 &
  4,517–11,590 &
  5,901 &
  12,836 &
  4,212 &
  16,980 &
  7,785 \\ \midrule
\textbf{OIL} &
  \begin{tabular}[c]{@{}c@{}}$K$: 5\\ $\gamma$: 0.5\\ $\alpha$: 0.99\\ $\beta$: 1\\ BATCH\_SIZE: 8\\ LR: 1e-6\end{tabular} &
  \begin{tabular}[c]{@{}c@{}}$K$: 5\\ $\gamma$: 0.5\\ $\alpha$: 0.99\\ $\beta$: 1\\ BATCH\_SIZE: 8\\ LR: 1e-6\end{tabular} &
  \begin{tabular}[c]{@{}c@{}}$K$: 3\\ $\gamma$: $\infty$\\ $\alpha$: 0.99\\ $\beta$: 1\\ BATCH\_SIZE: 16\\ LR: 1e-6\end{tabular} &
  \begin{tabular}[c]{@{}c@{}}$K$: 3\\ $\gamma$: $\infty$\\ $\alpha$: 0.99\\ $\beta$: 1\\ BATCH\_SIZE: 16\\ LR: 1e-6\end{tabular} &
  \begin{tabular}[c]{@{}c@{}}$K$: 3\\ $\gamma$: 0.5\\ $\alpha$: 0.99\\ $\beta$:1\\ BATCH\_SIZE: 16\\ LR: 1e-6\end{tabular} &
  \begin{tabular}[c]{@{}c@{}}$K$: 3\\ $\gamma$: $\infty$\\ $\alpha$: 0.99\\ $\beta$: 1\\ BATCH\_SIZE: 16\\ LR: 1e-6\end{tabular} &
  \begin{tabular}[c]{@{}c@{}}$K$: 1\\ $\gamma$: $\infty$\\ $\alpha$: 1\\ $\beta$:1\\ BATCH\_SIZE: 16\\ LR: 1e-6\end{tabular} &
  \begin{tabular}[c]{@{}c@{}}$K$: 1\\ $\gamma$: $\infty$\\ $\alpha$: 1\\ $\beta$: 1\\ BATCH\_SIZE: 16\\ LR: 1e-6\end{tabular} \\ \midrule
\textbf{PL} &
  \begin{tabular}[c]{@{}c@{}}$K$: 5\\ BATCH\_SIZE: 8\\ LR: 1e-6\end{tabular} &
  \begin{tabular}[c]{@{}c@{}}$K$: 5\\ BATCH\_SIZE: 8\\ LR: 1e-6\end{tabular} &
  \begin{tabular}[c]{@{}c@{}}$K$: 3\\ BATCH\_SIZE: 16\\ LR: 1e-6\end{tabular} &
  \begin{tabular}[c]{@{}c@{}}$K$: 1\\ BATCH\_SIZE: 16\\ LR: 1e-7\end{tabular} &
  \begin{tabular}[c]{@{}c@{}}$K$: 1\\ BATCH\_SIZE: 16\\ LR: 1e-7\end{tabular} &
  \begin{tabular}[c]{@{}c@{}}$K$: 1\\ BATCH\_SIZE: 16\\ LR: 1e-7\end{tabular} &
  \begin{tabular}[c]{@{}c@{}}$K$: 1\\ BATCH\_SIZE: 16\\ LR: 1e-7\end{tabular} &
  \begin{tabular}[c]{@{}c@{}}$K$: 1\\ BATCH\_SIZE: 16\\ LR: 1e-7\end{tabular} \\ \midrule
\textbf{Tent} &
  \begin{tabular}[c]{@{}c@{}}$K$: 5\\ BATCH\_SIZE: 8\\ LR: 1e-6\end{tabular} &
  \begin{tabular}[c]{@{}c@{}}$K$: 5\\ BATCH\_SIZE: 8\\ LR: 1e-6\end{tabular} &
  \begin{tabular}[c]{@{}c@{}}$K$: 3\\ BATCH\_SIZE: 16\\ LR: 1e-6\end{tabular} &
  \begin{tabular}[c]{@{}c@{}}$K$: 1\\ BATCH\_SIZE: 16\\ LR: 1e-7\end{tabular} &
  \begin{tabular}[c]{@{}c@{}}$K$: 1\\ BATCH\_SIZE: 16\\ LR: 1e-7\end{tabular} &
  \begin{tabular}[c]{@{}c@{}}$K$: 1\\ BATCH\_SIZE: 16\\ LR: 1e-7\end{tabular} &
  \begin{tabular}[c]{@{}c@{}}$K$: 1\\ BATCH\_SIZE: 16\\ LR: 1e-7\end{tabular} &
  \begin{tabular}[c]{@{}c@{}}$K$: 1\\ BATCH\_SIZE: 16\\ LR: 1e-7\end{tabular} \\ \bottomrule
\end{tabular}%
}
\caption{Hyper-parameters for TTA baselines. For MRQA, we find that for PL and OIL, when we keep the same learning rate and batch size as OIL, the final results are bad, so we choose better hyper-parameters for PL and Tent as the table shows.}
\label{tab:hyper-para}
\end{table*}

\subsection{Hyper-parameters}\label{app:hyper}

We provide the values of hyper-parameters for test-time adaptation. (1) For learning rate, we select a value smaller than the one used for training the source model. We set the learning rate to 1e-6. (2) For batch size, for smaller test sets, we set the batch size to 8. For larger test sets, we set the batch size to 16. (3) For $\alpha$ used in updating the expert model, if the test set is large, we set $\alpha$ to a larger value such as 1. Otherwise, we set $\alpha$ to a smaller value such as 0.99. (4) For $\gamma$ used in filtering the noisy labels, $\gamma = \infty$ works well for most of the test sets, except the datasets NoiseQA, XQuAD, and NaturalQA, where we set $\gamma$ to 0.5. (5) For memory size $K$, we set $K$ to a smaller value for large sets but to a larger value for small sets. The specific hyper-parameters used for TTA baselines are presented in Table~\ref{tab:hyper-para}.

\section{Effects of Denoising in OIL}
Table~\ref{tab:denoise} shows the effects of denoising in OIL. For NoiseQA and XQuAD, we set $\gamma$ to 0.5 to filter out the noisy labels. When using XLMR-large as the base model for NoiseQA-na, the average performance drops substantially if noisy labels are not removed.

\section{Results of RT methods on the Development Set of SQuAD}
Fig.~\ref{fig:gain-squad-dev} shows the gains on the development set of SQuAD trained by each RT baseline, to demonstrate the effectiveness of the RT methods. 

\section{Effects of Memory Size $K$}
Fig.~\ref{fig:memory-size} shows the effects of memory size $K$. We see that using a larger memory size can improve the adaptation results when the learning rate is not so large. When the learning rate is large, larger memory size can worsen the results. 

\end{document}